\documentclass{article}

\usepackage{arxiv}
\usepackage[utf8]{inputenc} 
\usepackage[T1]{fontenc}    
\usepackage{hyperref}       
\usepackage{url}            
\usepackage{booktabs}       
\usepackage{amsfonts}       
\usepackage{amsmath}       
\usepackage{nicefrac}       
\usepackage{microtype}      
\usepackage{lipsum}		
\usepackage{graphicx}
\usepackage{natbib}
\usepackage{doi}

\usepackage{tikz}
\usetikzlibrary{shapes}

\newcommand\figref{Figure~\ref}
\DeclareMathOperator*{\argmax}{\arg\!\max}

\title{Transforming Causality: Transformer-Based Temporal Causal Discovery with Prior Knowledge Integration}

\author{
    Jihua Huang \\ \texttt{jihua.huang@sri.com} \and 
    Yi Yao \\ \texttt{yi.yao@sri.com} \and 
    Ajay Divakaran \\ \texttt{ajay.divakaran@sri.com} 
}
\date{
    SRI International\\
    201 Washington Rd, Princeton, NJ 08540\\[2ex]
    \today
}

\renewcommand{\shorttitle}{\textit{arXiv} Template}

\hypersetup{
pdftitle={A template for the arxiv style},
pdfsubject={q-bio.NC, q-bio.QM},
pdfauthor={David S.~Hippocampus, Elias D.~Striatum},
pdfkeywords={First keyword, Second keyword, More},
}

\begin{document}
\maketitle

\begin{abstract}
    We introduce a novel framework for temporal causal discovery and inference that addresses two key challenges: complex nonlinear dependencies and spurious correlations. Our approach employs a multi-layer Transformer-based time-series forecaster to capture long-range, nonlinear temporal relationships among variables. After training, we extract the underlying causal structure and associated time lags from the forecaster using gradient-based analysis, enabling the construction of a causal graph. To mitigate the impact of spurious causal relationships, we introduce a prior knowledge integration mechanism based on attention masking, which consistently enforces user-excluded causal links across multiple Transformer layers. Extensive experiments show that our method significantly outperforms other state-of-the-art approaches, achieving a 12.8\% improvement in F1-score for causal discovery and 98.9\% accuracy in estimating causal lags.
\end{abstract}


\section{Introduction}\label{section:intro}

Understanding the causal mechanisms governing dynamic systems is essential for forecasting, decision-making, and scientific discovery \citep{prosperi2020causal, runge2019inferring, seth2015granger}. In many real-world domains, such as healthcare, finance, climate science, and neuroscience, data are usually represented as multivariate time series, where variables influence each other over time. In this setting, temporal causal discovery aims to uncover not only \textit{which} variables influence each other, but also \textit{when} those influences occur. Going further, causal inference enables reasoning under interventions -- asking what would happen if certain variables were manipulated. These tasks are foundational for interpretability and decision support in dynamic systems.

Despite their importance, temporal causal discovery and inference remain challenging due to nonlinear relationships, hidden confounders, lagged effects, and spurious correlations. Classical approaches such as constraint-based methods \citep{spirtes2000causation, runge2019detecting}, score-based optimization \citep{zheng2018dags, pamfil2020dynotears}, and Granger causality \citep{granger1969investigating} often rely on restrictive assumptions and do not scale well in high-dimensional or nonlinear settings.
To address these limitations, neural methods such as cMLP and cLSTM \citep{tank2021neural}, TCDF \citep{nauta2019causal}, DVGNN \citep{liang2024dynamic}, and CUTS \citep{cheng2023cuts} leverage deep architectures to improve modeling expressiveness. However, these models are typically built on recurrent or convolutional structures, which are limited in capturing long-range dependencies and global variable interactions.
In contrast, Transformer architectures, with their self-attention mechanisms, offer superior modeling of complex temporal structures and global dependencies. Transformer-based forecasting models like Informer \citep{zhou2021informer} and Autoformer \citep{wu2021autoformer} demonstrate strong performance in long-term time-series prediction, but they do not explicitly reason about causality and therefore cannot support interpretable or counterfactual analysis.
The most directly related method, CausalFormer \citep{kong2024causalformer}, introduces a Transformer-based approach for causal structure learning. However, it is restricted to single-layer architecture, limiting its ability to capture complex nonlinear causal dependencies or accommodate very long temporal sequences.

In this work, we propose a novel framework for temporal causal discovery and inference, composed of a causality-aware time-series forecaster and an architecture-agnostic causal graph extractor. The forecaster employs a multi-layer Transformer-based architecture to capture complex nonlinear and long-range temporal dependencies among time-series variables during training. The multi-layer design further enables the model to capture multi-resolution temporal features, enhancing its ability to represent both short- and long-term causal influences. Once these causal relationships are implicitly encoded within the forecaster, the causal graph extractor identifies causal links and their corresponding time lags by analyzing the model's gradients, ultimately constructing a directed causal graph. To integrate human domain knowledge with the learned causal structure, we introduce an attention masking based prior knowledge integration mechanism, which enforces user-specified exclusions of causal links consistently across all Transformer layers. This enables a human-in-the-loop refinement process, allowing domain experts to iteratively prune spurious causal relationships and improve the reliability of the resulting model for causal inference.

In summary, our main contributions are as follows:
\begin{enumerate}
    \item[\textbullet] We propose a novel framework for temporal causal discovery and inference that employs a multi-layer Transformer architecture to model complex nonlinear relationships and long-range temporal dependencies in time-series data.
    \item[\textbullet] We design an architecture-agnostic, gradient-based module for causal graph extraction that interprets a trained time-series forecaster to recover causal graphs and associated time lags. Its model-agnostic nature ensures broad applicability across diverse model architectures.
    \item[\textbullet] We introduce a prior knowledge integration mechanism based on attention masking, which consistently enforces user-specified causal constraints across all Transformer layers. This design enables our framework to exploit a multi-layer architecture necessary for modeling complex nonlinear dependencies. It also allows for human-in-the-loop refinement to prevent causal inference from being misled by spurious correlations in data.
    \item[\textbullet] We conduct extensive experiments demonstrating that our method substantially outperforms other state-of-the-art approaches, achieving a 12.8\% improvement in F1-score for causal discovery and 98.9\% accuracy in estimating causal lags.
\end{enumerate}

\section{Related Work}\label{section:related}

\subsection{Classical Methods for Temporal Causal Discovery}
Causal discovery in time-series data is a long-standing problem that aims to recover both the causal graph and the corresponding time lags among variables. Classical approaches can be broadly grouped into constraint-based, score-based, and Granger causality frameworks.

Constraint-based methods such as the PC algorithm and FCI \citep{spirtes2000causation} identify causal relations by testing conditional independencies. PCMCI and PCMCI+ \citep{runge2019detecting, gerhardus2020high} extend these methods to time-series by handling temporal autocorrelation and offering improved scalability with nonlinear independence tests.

Score-based approaches optimize a global objective, often involving penalized likelihoods or structural constraints. Examples include NOTEARS \citep{zheng2018dags} and its dynamic extension DYNOTEARS \citep{pamfil2020dynotears}, which enforce acyclicity using continuous optimization.

Granger causality \citep{granger1969investigating} remains widely used in econometrics and neuroscience. It tests whether past values of one variable help predict another, typically using linear vector autoregressive (VAR) models. While popular, standard Granger causality fails in nonlinear or confounded settings. Nonlinear extensions have been proposed \citep{marinazzo2008kernel}, but they generally lack identifiability guarantees.

These classical approaches laid the foundation for causal inference in time-series, but often struggle to capture complex nonlinear dependencies and offer limited support for high-dimensional data.

\subsection{Deep Learning-Based Causal Discovery}
Recent years have seen a surge of interest in deep learning methods for causal discovery, which offer improved flexibility and scalability.

cMLP and cLSTM \citep{tank2021neural} use sparse feedforward and recurrent networks, respectively, to predict each variable from its potential causes. By applying sparsity-inducing penalties, these methods infer causal graphs from learned connectivity patterns. TCDF \citep{nauta2019causal} extends this idea with dilated CNNs and gradient-based attention to recover both structure and causal lags.

More recently, CausalFormer \citep{kong2024causalformer} introduces Transformers into causal discovery, using attention weights and structural masks to uncover time-lagged causal relationships. However, it is limited to single-layer architectures and lacks support for enforcing prior constraints or human input.

Graph-based models such as DVGNN \citep{liang2024dynamic} and CUTS \citep{cheng2023cuts} leverage dynamic graph neural networks to learn time-evolving causal graphs. These models can capture rich temporal dependencies but often rely on latent representations that hinder interpretability and constraint enforcement.

Our approach differs in that we leverage a multi-layer Transformer architecture, enabling deeper modeling of nonlinear interactions, while enforcing cross-layer causal consistency and supporting prior knowledge injection via attention masking.

\subsection{Prior Knowledge Integration in Causal Discovery}
Incorporating domain knowledge into causal discovery is essential when observational data alone is insufficient to resolve ambiguities, especially under confounding or noise. Prior knowledge can take the form of excluded edges or temporal ordering constraints.

Constraint-based methods often support static incorporation of prior knowledge via edge restrictions \citep{runge2019detecting}. Score-based methods such as NOTEARS \citep{zheng2018dags} and DYNOTEARS \citep{pamfil2020dynotears} also allow for partial constraint specification. However, these constraints are typically applied only at initialization or as penalties in the optimization objective, without persistent enforcement in deep models.

In contrast, our framework provides a mechanism for persistent constraint enforcement through attention masking applied consistently across all Transformer layers. This allows users to specify excluded causal links, which are strictly respected during both training and inference. Moreover, it supports a human-in-the-loop refinement process, enabling users to iteratively prune spurious causal links identified by the model, thereby improving robustness and interpretability.

\begin{figure*}[!t]
    \includegraphics[clip, trim=1.7cm 0.95cm 1.7cm 0.77cm, width=\textwidth]{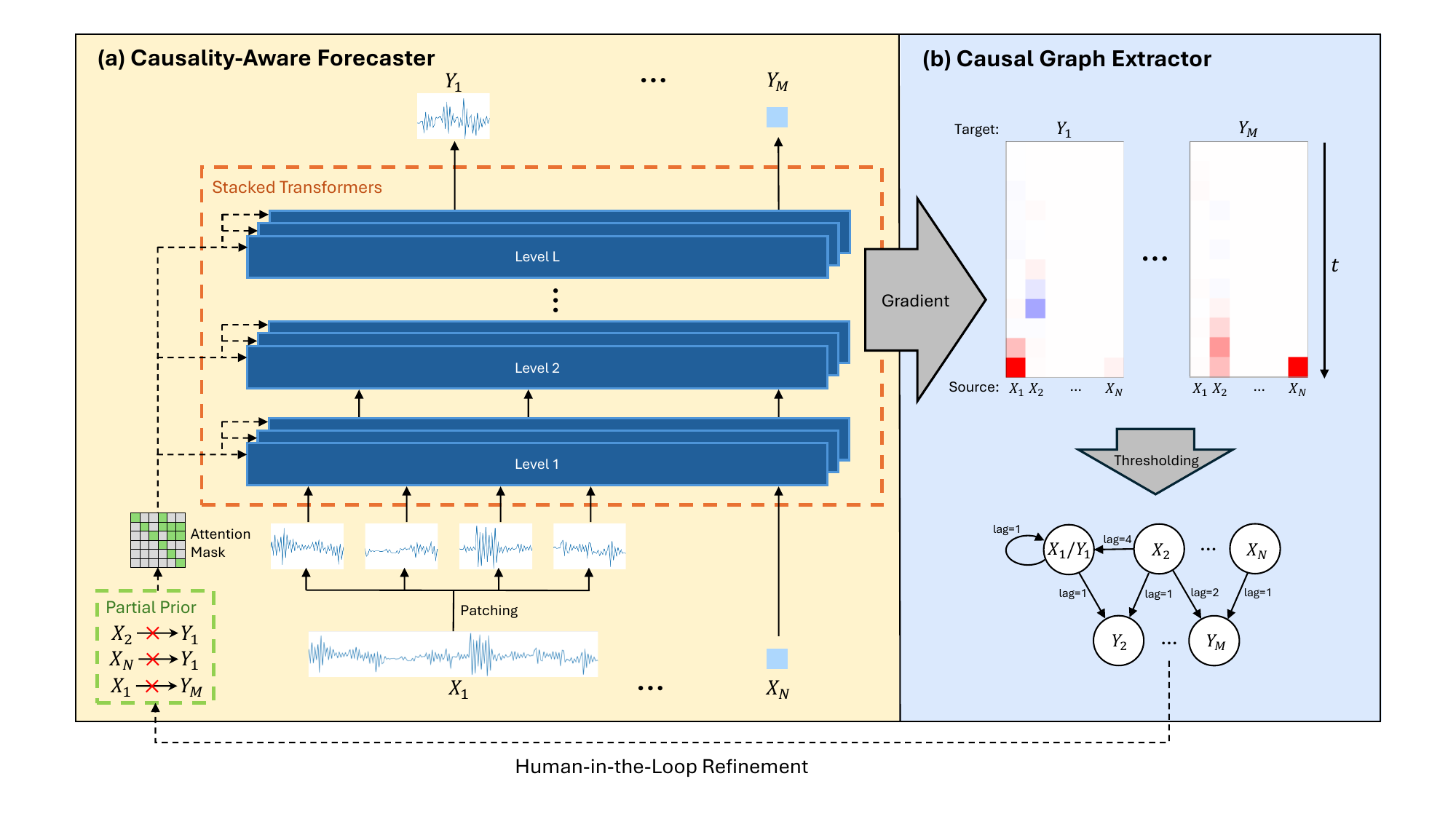}
    \caption{\textbf{Overview of the proposed causal discovery and inference framework}. The framework consists of two key components. (a) The causality-aware forecaster learns to predict target variables from input variables, while implicitly capturing causal dependencies during the training process. It also incorporates prior knowledge, when available, by applying attention masks to explicitly constrain the variable dependencies. (b) The causal graph extractor computes the gradients of the forecaster with respect to the target variables, extracting causal links and constructing a corresponding causal graph. The framework supports iterative human-in-the-loop refinement, enabling users to modify the generated causal graph and integrate these modifications as prior knowledge to further refine the forecaster.}
    \label{fig:overview}
\end{figure*}


\section{Methods}\label{section:methods}

Causal discovery aims to identify causal relationships among a set of variables $U$ and represent these relationships through a causal graph $G=(U,E)$ where $E$ is the set of causal links between pairs of variables. Each causal link originated from a time-series variable has an associated causal lag. 

In many practical applications, the focus is often on identifying causal chains leading to a small number of key variables, such as product defect rate. We refer to these key variables as target variables $Y$, and their potential causes as source variables $X$. It is important to note that $X$ and $Y$ are not mutually exclusive, as a key variable can be the cause of another. When $X$ and $Y$ both encompass all variables in $U$, the goal is to identify causal relationships among all variables in the system. 
This distinction between source and target variables not only directs model focus to the causal relationships relevant to the target key variables, but also enables prior knowledge integration and human-in-the-loop refinement. Further details are provided in Section \ref{section:prior}.

Let $X = \left[X_1, X_2, \dots, X_N \right]$ be the $N$ source variables and $Y = \left[Y_1, Y_2, \dots, Y_M\right]$ be the $M$ target variables. Each variable $X_i$ (and $Y_j$) belongs to one of the four data types:
\begin{enumerate}
    \item Static numerical: $X_i \in \mathbb{R}^1$; 
    \item Static categorical: $X_i \in \{1,2,\cdots,C_i\}^1$ where $C_i$ is the number of unique categories; 
    \item Time-series numerical: $X_i \in \mathbb{R}^L$ where $L$ is the total number of samples; 
    \item Time-series categorical: $X_i \in \{1,2,\cdots,C_i\}^L$. 
\end{enumerate}
These variables may contain missing values, indicated by NaN. We resample the source time-series variables by interpolation so that they all have the same sampling rate and the number of samples $S$.

\figref{fig:overview} illustrates the workflow of the proposed causal discovery and inference framework, which consists of two main components: the causality-aware forecaster and the causal graph extractor. 
First, the causality-aware forecaster (\figref{fig:overview}a) is trained to predict future values of the target variables based on the history of the source variables. Causal relationships are implicitly captured during this process, as they are necessary for accurate forecasting. When partial prior knowledge is available, excluded causal links are specified and enforced using attention masks, preventing the forecaster from learning spurious correlations from the data. 
Next, the causal graph extractor (\figref{fig:overview}b) computes the gradients for each target variable, extracts causal relationships and lags from large gradients, and constructs a causal graph accordingly.
Finally, if the generated causal diagram contains spurious correlations, users can manually remove the corresponding edges and re-run the process to iteratively incorporate human knowledge and refine the forecaster as well as the causal graph.

\subsection{Causality-Aware Forecaster}

The Causality-Aware Forecaster is built upon a multi-layer Transformer-based architecture, designed to effectively capture nonlinear dependencies among time-series variables. To handle long input sequences where the number of patches exceeds the capacity of a single Transformer, we employ a stacked Transformer design that performs convolution along the temporal dimension. This enables hierarchical extraction of multi-resolution temporal features, allowing the model to represent both short- and long-term dependencies. Throughout this process, prior causal constraints are incorporated via attention masks, allowing the model to seamlessly integrate human expertise with the causal relationships it learns from the data.

\subsubsection{Input Tokenization}

\textbf{Normalization} Numerical variables are linearly scaled such that the range between the 5th and 95th percentiles is mapped to the interval [0, 1].

\textbf{Patching} Instead of treating each sample of time-series variables as a single token, \citep{nie2022time} demonstrates that partitioning time-series data into patches and then tokenizing these patches leads to significant improvements in forecasting accuracy for transformer-based models. Therefore we divide each input univariate time series variable of length $S$ into $T$ patches of length $P$ with stride $S_\text{patch}$, where $T = \lfloor (S-P) / S_\text{patch} \rfloor + 1$.


\textbf{Encoding} Each variable $X_i$ is encoded into tokens $x_i$ based on its data type:
\begin{enumerate}
    \item Static numerical: Through a trainable linear projection $W^\text{encode}_i \in \mathbb{R}^D$, where $D$ is the embedding size.
    \item Static categorical: Through an embedding encoder parameterized by a trainable weight matrix $W^\text{encode}_i \in \mathbb{R}^{D \times C_i}$ where $C_i$ is the number of unique categories.
    \item Time-series numerical: Each patch is encoded through a trainable linear projection $W^\text{encode}_i \in \mathbb{R}^{D \times P}$, where $P$ is the patch size.
    \item Time-series categorical: First, each sample in the patch is embedded through an embedding encoder with weight $W^\text{encode}_{i,1} \in \mathbb{R}^{D' \times C_j}$, where $D'$ is the intermediate embedding size typically much smaller than $D$. Second, the $P$ embedded samples in this patch are concatenated and linearly mapped to a latent representation through a trainable linear projection with weight $W^\text{encode}_{i,2} \in \mathbb{R}^{D \times D' \cdot P}$.
\end{enumerate}

Embeddings corresponding to missing values or patches are replaced by learnable special tokens. A distinct missing value token $W^{\text{miss}}_i \in \mathbb{R}^{D}$ is learned for each variable $X_i$ (and for $Y_j$ as well).

After patching and encoding, each static variable $X_i$ is embedded as a single token $x_i \in \mathbb{R}^{D}$, while each time-series variable $X_j$ is embedded as $T$ tokens $x_j = [x_j^1, x_j^2, \cdots, x_j^T]$. The number of tokens $T = |x_j|$ is the same as the number of patches in $X_j$.

\subsubsection{Stacked Transformers}

The embedded tokens are processed through a stack of $L$ Transformer levels, denoted as $\mathbf{F} = [\mathbf{F}^1,\mathbf{F}^2,\cdots,\mathbf{F}^L]$, and the transformation at each level $l$ is:
\begin{equation}
    y_0^{l}, \cdots, y_M^{l}, x_0^{l}, \cdots, x_N^{l} = \mathbf{F}^l(y_0^{l-1}, \cdots, y_M^{l-1}, x_0^{l-1}, \cdots, x_N^{l-1})
\end{equation}
Each Transformer level $\mathbf{F}^l$ consists of $K$ transformer layers.
In the first transformer level $\mathbf{F}^1$, the input tokens $y_j^{0} \in \mathbb{R}^{D}$ are trainable sentinel tokens, similar to the classification token in transformer-based classifiers.

To aggregate information in long time-series data, each transformer level $\mathbf{F}^l$ reduces the length of time-series tokens by performing "1D convolution" with window size $W_{\text{conv}}$ and stride $S_{\text{conv}}$ along the temporal dimension. Each sliding window processes one token per static variable and $W_{\text{conv}}$ tokens per time-series variable, and outputs one token for each static and time-series variable by averaging the output tokens corresponding to the time-series variables. 
Let $T^{l-1}$ be the number of tokens for the time-series variables before transformer level $l$, then the number of sliding windows is
\begin{equation}
    T^l = \lceil \frac{T^{l-1} - W_{\text{conv}}}{S_{\text{conv}}} \rceil + 1
\end{equation}
After transformer level $l$, each time-series variable has $T^l$ tokens, one from each sliding window. For static variables, their $T^l$ tokens are averaged and reduced to one token, as static variables do not need a time-varying embedding.

The total number of stacked transformer level $l$ are automatically determined as the minimum number of levels reducing the number of time-series tokens to 1: $T^L = 1$.

\subsubsection{Output Decoding}

Each token $y_j^L \in \mathbb{R}^D$ is decoded to forecast output $\hat{Y}_j$ based on its data type:
\begin{enumerate}
    \item Static numerical: through a linear projection $W^\text{decode}_j \in \mathbb{R}^D$, yielding $\hat{Y}_j \in \mathbb{R}^1$.
    \item Static categorical: through a linear projection $W^\text{decode}_j \in \mathbb{R}^{C_j \times D}$ where $C_j$ is the number of unique categories, yielding $\hat{Y}_j \in \mathbb{R}^{C_j}$, a vector of logits corresponding to each category.
    \item Time-series numerical: through a linear projection $W^\text{decode}_j \in \mathbb{R}^{S_j \times D}$ where $S_j$ is the forecast window size, yielding $\hat{Y}_j \in \mathbb{R}^{S_j}$.
    \item Time-series categorical: through a linear projection $W^\text{decode}_j \in \mathbb{R}^{S_j \times C_j \times D}$, yielding $\hat{Y}_j \in \mathbb{R}^{S_j \times C_j}$.
\end{enumerate}

To train the forecaster, we apply L1 loss for the numerical variables and cross entropy loss for the categorical variables:
\begin{equation}
    \mathcal{L}(\hat{Y}, Y) = \sum_{i \in \mathcal{N}} \left| y_i - \hat{y}_i \right| + \sum_{j \in \mathcal{C}} - \log \left( \frac{\exp(\hat{y}_{j,Y_j})}{\sum_{c=1}^{C_j} \exp(\hat{y}_{j,c})} \right)
\end{equation}
where $\mathcal{N}$ denotes the set of indices for numerical variables, and $\mathcal{C}$ denotes the set of indices for categorical variables.

\subsubsection{Incorporation of Prior Knowledge}\label{section:prior}

\begin{figure*}[!b]
    \centering\includegraphics[clip, trim=0cm 0cm 0cm 0cm, width=0.75\textwidth]{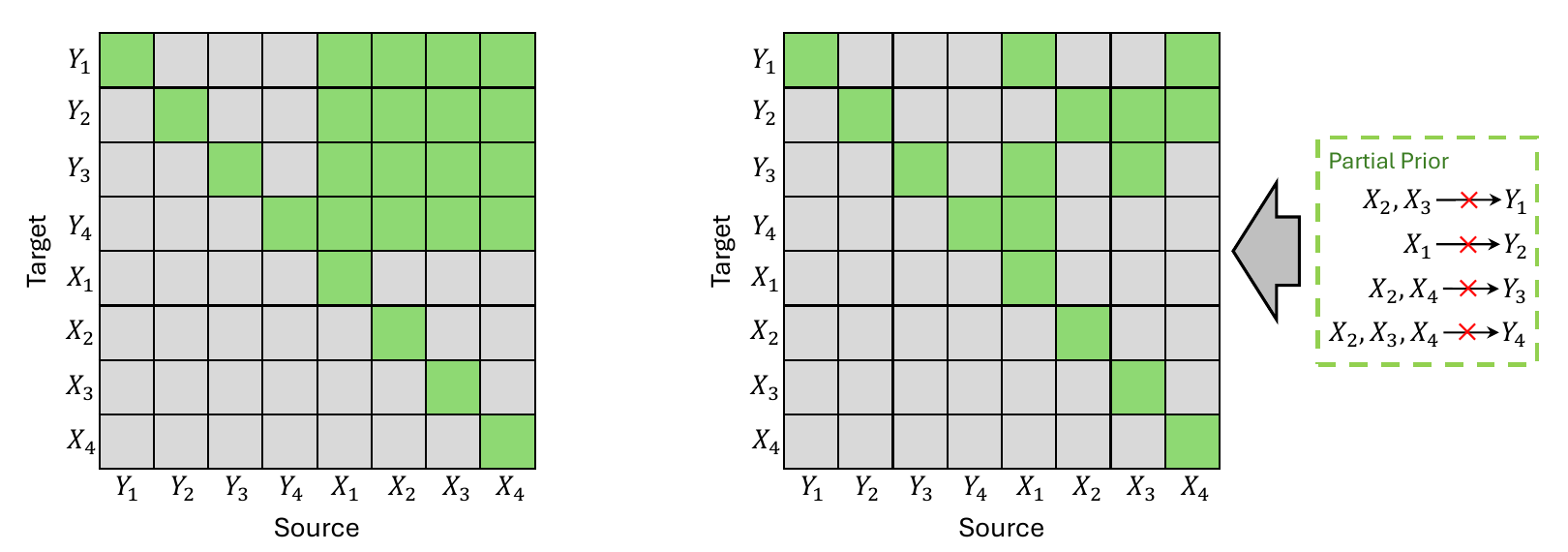}
    \put(-285,-6){\footnotesize \textbf{(a)} Default attention mask}
    \put(-155,-6){\footnotesize \textbf{(b)} Attention mask based on prior knowledge}
    \caption{\textbf{Attention Mask Examples}. (a) Default attention mask when no prior knowledge is available. (b) Attention mask derived from the provided excluded causal links. The masks are binary matrices, where gray entries correspond to 0 and green entries to 1.}
    \label{fig:attention_mask}
\end{figure*}

Human domain knowledge plays a crucial role in complementing limited training data and in pruning spurious correlations, such as higher ice cream sales causing increased drowning incidents. When partial prior knowledge is available in the form of excluded causal links, the proposed framework incorporates these constraints using attention masks (\figref{fig:attention_mask}). The attention mask $A \in \mathbb{B}^{(M+N) \times (M+N)}$ is a binary matrix, where each entry $A[i,j]$ indicates whether the $j$-th variable is a valid cause of the $i$-th variable. 

In the absence of prior knowledge, the attention mask is constructed based on the definition of source and target variables. Specifically, all source variables $X$ are allowed to influence the target variables $Y$, and no causal links are permitted within either the source or target variables themselves (\figref{fig:attention_mask}a).

When excluded causal links are provided, the corresponding entries in the attention mask are set to zero (\figref{fig:attention_mask}b), effectively disabling these relationships during attention computation.

The attention mask is applied at each layer of the stacked Transformers. Since some variables are time-series variables represented by multiple tokens, we replicate their corresponding rows and columns of the attention mask $W_{\text{conv}}$ times to match the total number of tokens processed by each sliding window. As a result, the attention mask for the $l$-th Transformer level, denoted $A^l$, has the same dimensions as the key and query matrices it operates on.

The attention for the $l$-th Transformer level is then computed as
\begin{equation}
    \text{Attention}^l(Q, K, V) = \text{softmax}\left( \frac{QK^T}{\sqrt{D}} \cdot A^l \right) V
\end{equation}

This attention mask ensures that variables receive zero attention from their excluded sources, strictly enforcing the provided prior knowledge during the attention computation in each transformer layer.

While the attention masks enforce prior knowledge at the level of individual transformer layers, global enforcement requires preventing the propagation of attention across multiple layers. We achieve this global enforcement by source and target variable separation. To illustrate, consider a simple example involving three variables: $U = [U_1, U_2, U_3]$, and a prior constraint stating that $U_1$ should not cause $U_3$. If we do not separate the source and target groups (into $X = [X_1, X_2, X_3]$ and $Y = [Y_1, Y_2, Y_3]$) and instead use an attention mask $\tilde{A} \in \mathbb{R}^{3 \times 3}$ (as opposed to $\mathbb{R}^{6 \times 6}$), then after merely two transformer layers, token $u^2_3$ would no longer be independent of $u^0_1$. This is because $u^0_1$ influences $u^1_2$, which subsequently affects $u^2_3$, violating the prior constraint. In contrast, our proposed method prevents such attention propagation: since source and target variables are kept distinct, $y^l_3$ remains unaffected by $x^k_1$ or its dependents $y^{l-1}_1, y^{l-1}_2$ in every layer $l$, thereby ensuring that unintended attention propagation is prevented across all layers.

\subsection{Causal Diagram Extraction}

\begin{figure*}[!b]
    \includegraphics[clip, trim=0cm 2cm 13cm 2cm, width=\textwidth]{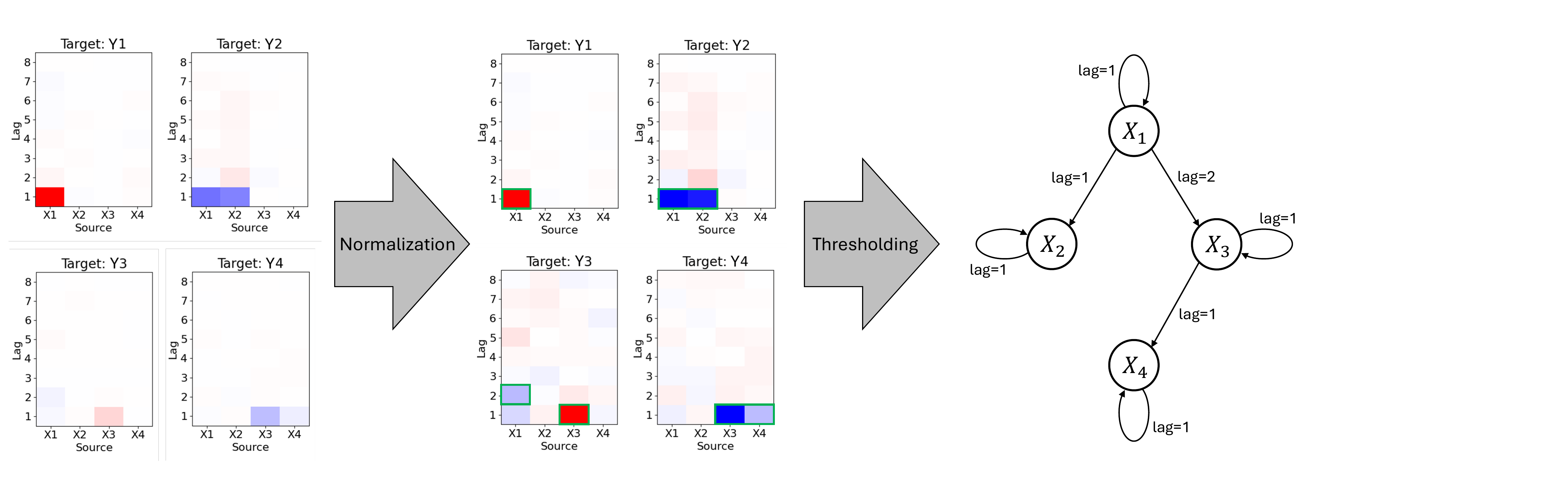}    
    \put(-390,-12){\footnotesize \textbf{(a)} Forecaster Gradients}
    \put(-252,-12){\footnotesize \textbf{(b)} Normalized Gradients}
    \put(-102,-12){\footnotesize \textbf{(c)} Extracted Causal Graph}
    \caption{\textbf{Causal Diagram Extraction}. This example includes four time-series variables, each treated as both source and target variables. (a) Forecaster gradient $\nabla\mathbf{F}$ for each target variable. Color intensity indicates gradient magnitude, with red representing positive values and blue representing negative values. (b) Normalized causal scores $\nabla\mathbf{F}^\text{norm}$, with entries used to determine causal lags highlighted by a green rectangle. (c) The resulting causal graph $G$.}
    \label{fig:causal_diagram_extraction}
\end{figure*}

To construct the causal diagram, we extract causal links from the trained forecaster $\mathbf{F}$ by computing the gradient for each target variable. Let $\mathbf{F}_j(X_1, \cdots, X_N)$ denote the $j$-th output of the forecaster which is $Y_j$. Its gradient $\nabla\mathbf{F}_j \in \mathbb{R}^{S \times N}$ with respect to all $N$ input variables is computed as follows:
\begin{equation}
    \nabla\mathbf{F}_j(s,i) = \begin{cases}
        \mathbf{F}_j(\cdots, X_i + \epsilon, \cdots) - \mathbf{F}_j(\cdots, X_i - \epsilon, \cdots) & \text{if } X_i \text{ is static and } s=1\\
        0 & \text{if } X_i \text{ is static and } 2 \le s \le S\\
        \mathbf{F}_j(\cdots, X_i + \epsilon \cdot e_s, \cdots) - \mathbf{F}_j(\cdots, X_i - \epsilon \cdot e_s, \cdots) & \text{if } X_i \text{ is a time-series}
    \end{cases}
\end{equation}
where $\epsilon \in (0,1]$ is a small delta added to the normalized input $X_i$, $e_s \in \mathbb{R}^S$ is a unit vector non-zero only in the $s$-th entry, and S is the number of samples in the time-seris variables. An example of the computed gradients is shown in \figref{fig:causal_diagram_extraction}a.

Next, we normalize each gradient $\nabla\mathbf{F}_j$ by dividing its maximum value to the obtain causal scores $\nabla\mathbf{F}^\text{norm}_j$:
\begin{equation}
    \nabla\mathbf{F}^\text{norm}_j = \dfrac{\nabla\mathbf{F}_j}{\max_{s,i} \nabla\mathbf{F}_j(s,i)}
\end{equation}
This normalization ensures that the strongest causal effect is assigned a score of 1, allowing us to filter out insignificant causes by applying a user-defined threshold $\tau \in [0,1]$ to their relative causal scores (\figref{fig:causal_diagram_extraction}b).

Based on $\nabla\mathbf{F}^\text{norm}$, the causal graph $G$ is constructed by adding a directed edge $X_i \xrightarrow{} Y_j$ if the maximum causal score from $X_i$ to $Y_j$ over all time lags is no less than $\tau$:
\begin{equation}
    \max_{s=1}^S |\nabla \mathbf{F}^\text{norm}_j(s,i)| \ge \tau
\end{equation}
In the case where $X_i$ represents a time-series variable, the causal lag associated with this edge is determined by:
\begin{equation}
    \text{Lag}(X_i, Y_j) = \argmax_{s=0}^S |\nabla \mathbf{F}^\text{norm}_j(s,i)|
\end{equation}
An example of the constructed causal diagram is illustrated in \figref{fig:causal_diagram_extraction}c.

\section{Experiments}\label{section:experiments}

\subsection{Datasets}\label{section:datasets}
We evaluate our method on three datasets: a synthetic dataset modeling basic causal structures \citep{assaad2022entropy}, a simulated nonlinear climate dataset \citep{lorenz1996predictability}, and a real-scenario inspired brain network simulation dataset \citep{smith2011network}.

\begin{figure*}[!htb]
    \includegraphics[clip, trim=0cm 8cm 6cm 0cm, width=\textwidth]{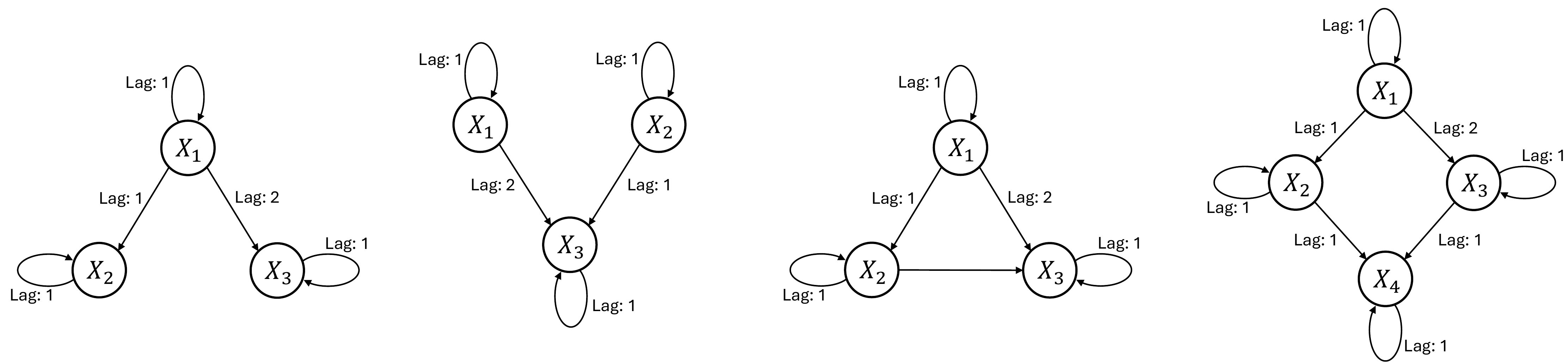}    
    \put(-365,-10){\footnotesize \textbf{(a)} Fork}
    \put(-277,-10){\footnotesize \textbf{(b)} V-structure}
    \put(-177,-10){\footnotesize \textbf{(c)} Mediator}
    \put(-70,-10){\footnotesize \textbf{(c)} Diamond}
    \caption{\textbf{Four basic causal structures in the synthetic dataset}.}
    \label{fig:dataset_Harvard_synthetic}
\end{figure*}
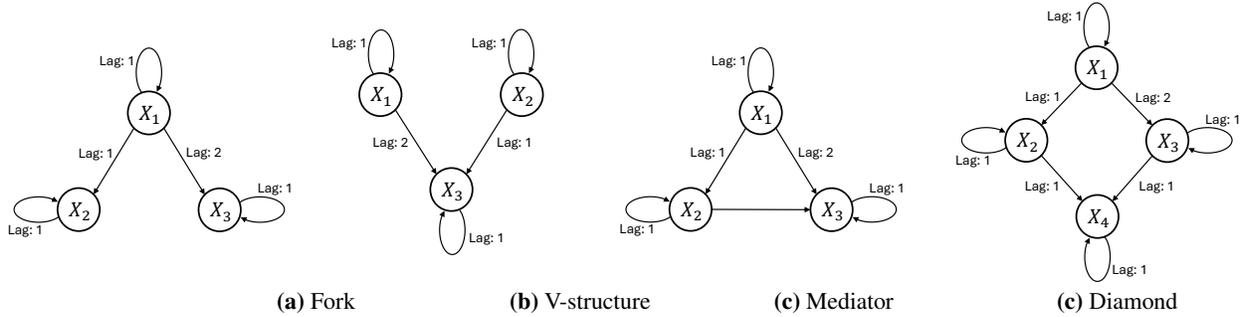

\textbf{Basic Structures} This synthetic dataset resembles four basic causal structures with additive random noise: fork, V-structure, mediator, and diamond (\figref{fig:dataset_Harvard_synthetic}). 
For each structure, 10 causal graphs were created, each with random functions and coefficients assigned to its edges. From each causal graph, three time-series of 1000 samples were generated, with random Gaussian noise added to the data.

\textbf{Lorenz96} This dataset is simulated using the nonlinear Lorenz-96 model of climate dynamics:
\begin{equation}
    \frac{dX_i^t}{dt} = (X_{i+1}^t - X_{i-2}^t) \cdot X_{i-1}^t - X_i^t + F
\end{equation}
where $X_i^t$ denotes the value of time series variable $X_i$ at time t, and a constant $F \in [30,40]$ controls the level of chaotic forcing. Each simulation includes 10 time-series variables of length 1000.

\textbf{NetSim} This dataset contains realistic simulations of functional MRI (fMRI) data, designed for benchmarking brain network estimation methods. It includes time-series data from 28 diverse brain networks with 5, 10, 15 or 50 nodes, spanning a range of network densities and topologies. Its time series lengths vary from 50 to 5000 timepoints.

\subsection{Evaluation Metrics}
We evaluate the performance of causal discovery methods with the two metrics below.

\textbf{F1-score}: computed from precision and recall on the edges in the reconstructed causal diagram.

\textbf{Precision of delay (PoD)}: mean average error (MAE) on the estimated delay associated to edges in the causal diagram whose source node is a time-series variable. 

For each metric, we report the mean and standard deviation across the multiple time-series in the dataset.

\subsection{Implementation Details}\label{section:implementation}
All variables in the three datasets are treated as both source variables and target variables. Each source time-series variable contains $S=8$ samples and target variable forecast window size is set to $S_j = 1$ sample. We use a patching size of $P = 8$ and a stride of $S_\text{patch} = 8$, meaning each input time-series contains a single patch of 8 samples. The stacked Transformers architecture consists of $L=1$ level of $K=4$ Transformer layers, each having 8 attention heads and a hidden dimension of $D=128$. The forecaster is optimized using the Adam optimizer with a learning rate of $10^{-3}$ for 500 epochs. For causal graph extraction, gradients are computed with perturbation $\epsilon=0.05$. These gradients are averaged over all training samples, and a threshold $\tau=0.15$ is applied to identify causal links and corresponding time lags. All models are trained and evaluated on an NVIDIA GeForce GTX 1080 GPU with 8GB of memory, coupled with an Intel(R) Xeon(R) E5-2687W v4 CPU running at 3GHz.

\begin{table}[!t]
    \centering
    \resizebox{\linewidth}{!}{
        \begin{tabular}{ l|ccccccc } 
            \hline
                                        & cMLP & cLSTM & TCDF & DVGNN & CUTS & CausalFormer & Ours \\ 
            \hline
            Basic Structures            & 0.63±0.20 & 0.57±0.22 & 0.72±0.09 & 0.70±0.04 & 0.50±0.19 & 0.74±0.08 & \textbf{0.83±0.10} \\
            \hspace{6.5mm}Fork          & 0.51±0.33 & 0.47±0.32 & 0.73±0.10 & 0.75±0.00 & 0.50±0.19 & 0.79±0.11 & \textbf{0.87±0.07} \\
            \hspace{6.5mm}V-structure   & 0.73±0.15 & 0.60±0.20 & 0.76±0.09 & 0.73±0.06 & 0.49±0.15 & 0.77±0.05 & \textbf{0.83±0.13} \\
            \hspace{6.5mm}Mediator      & 0.71±0.14 & 0.59±0.24 & 0.69±0.06 & 0.65±0.05 & 0.52±0.23 & 0.71±0.06 & \textbf{0.83±0.11} \\
            \hspace{6.5mm}Diamond       & 0.55±0.19 & 0.63±0.13 & 0.68±0.09 & 0.65±0.04 & 0.49±0.20 & 0.68±0.08 & \textbf{0.80±0.09} \\
            Lorenz96                    & 0.64±0.03 & 0.63±0.06 & 0.46±0.05 & 0.48±0.07 & 0.58±0.02 & 0.69±0.06 & \textbf{0.84±0.03} \\
            NetSim                      & 0.58±0.14 & 0.56±0.13 & 0.59±0.12 & 0.56±0.12 & 0.61±0.13 & 0.66±0.09 & \textbf{0.80±0.09} \\
            \hline
        \end{tabular}
    }
    \caption{F1-score (mean ± std) for causal discovery methods. Highest scores shown in bold.}
    \label{table:eval_F1_Harvard_synthetic}
\end{table}

\subsection{Results}
Table \ref{table:eval_F1_Harvard_synthetic} reports the F1-scores (mean ± standard deviation) for causal graph reconstruction across the proposed method and several state-of-the-art (SOTA) baselines. The proposed approach consistently achieves superior performance across all three datasets, with F1-score improvements of 9.4\% on the Basic Structures dataset, 15\% on the Lorenz96 dataset, and 14\% on the NetSim dataset.

Table \ref{table:eval_PoD_Harvard_synthetic} presents the Precision of delay (PoD) for the proposed method and the SOTA methods. The proposed method accurately estimates causal lags for 97.8\%, 99\%, and 100\% of the causal diagram edges in the respective datasets, outperforming all SOTA baselines in this metric as well.

\begin{table}[!t]
    \centering
    \begin{tabular}{ l|cccc } 
        \hline
                                    & cMLP & TCDF & CausalFormer & Ours \\ 
        \hline
        Basic Structures            & 0.85±0.22 & 0.97±0.06 & 0.61±0.33 & \textbf{0.98±0.06} \\
        \hspace{6.5mm}Fork          & 0.76±0.41 & \textbf{1.00±0.00} & 0.46±0.34 & 0.97±0.07 \\
        \hspace{6.5mm}V-structure   & 0.91±0.16 & \textbf{1.00±0.00} & 0.59±0.39 & \textbf{1.00±0.00} \\
        \hspace{6.5mm}Mediator      & 0.91±0.12 & \textbf{0.97±0.11} & 0.63±0.40 & \textbf{0.97±0.07} \\
        \hspace{6.5mm}Diamond       & 0.82±0.17 & 0.92±0.13 & 0.74±0.20 & \textbf{0.97±0.10} \\
        Lorenz96                    & 0.45±0.17 & 0.77±0.08 & 0.42±0.18 & \textbf{0.99±0.01} \\
        NetSim                      &     -     &      -    & 0.41±0.37 & \textbf{1.00±0.02} \\
        \hline
    \end{tabular}
    \caption{PoD (mean ± std) for causal discovery methods. Highest scores shown in bold.}
    \label{table:eval_PoD_Harvard_synthetic}
\end{table}

\section{Conclusion}\label{section:conclusion}

We presented a novel framework for temporal causal discovery and inference that leverages a multi-layer Transformer-based forecaster to capture complex nonlinear and long-range temporal dependencies. Causal graphs are subsequently constructed through gradient-based analysis of the trained forecaster. In addition, we introduced a prior knowledge integration mechanism using attention masking, enabling users to prune spurious causal relationships by specifying causal constraints.

Our work directly addresses two central challenges in temporal causal discovery: modeling nonlinear dependencies and mitigating spurious correlations. Beyond these, several important challenges remain, including handling irregularly sampled time series and modeling context-specific causal relationships. Future work may explore continuous-time Transformer architectures such as \citep{chen2023contiformer} to accommodate irregular sampling. Addressing context-dependent causal structures may require moving beyond causal graphs representation —- an open and promising direction for advancing the field.

\bibliographystyle{unsrtnat}
\bibliography{arxiv} 

\end{document}